\documentclass[sigconf]{acmart}

\renewcommand\footnotetextcopyrightpermission[1]{} 
\setcopyright{none}

\begin{document}

\title{Kinematics-based 3D Human-Object Interaction Reconstruction from Single View}

\author{Yuhang Chen}
\affiliation{%
  \institution{School of Automation, Southeast University}
  \city{Nanjing}
  \country{China}
}
\email{chenyuhang@seu.edu.cn}

\author{Chenxing Wang}
\authornote{Corresponding Author}
\affiliation{%
  \institution{School of Automation, Southeast University}
  \city{Nanjing}
  \country{China}
}
\email{cxwang@seu.edu.cn}



\begin{abstract}
Reconstructing 3D human-object interaction (HOI) from single-view RGB images is challenging due to the absence of depth information and potential occlusions. Existing methods simply predict the body poses merely rely on network training on some indoor datasets, which cannot guarantee the rationality of the results if some body parts are invisible due to occlusions that appear easily. Inspired by the end-effector localization task in robotics, we propose a kinematics-based method that can drive the joints of human body to the human-object contact regions accurately. After an improved forward kinematics algorithm is proposed, the Multi-Layer Perceptron is introduced into the solution of inverse kinematics process to determine the poses of joints, which achieves precise results than the commonly-used numerical methods in robotics. Besides, a Contact Region Recognition Network (CRRNet) is also proposed to robustly determine the contact regions using a single-view video. Experimental results demonstrate that our method outperforms the state-of-the-art on benchmark BEHAVE. Additionally, our approach shows good portability and can be seamlessly integrated into other methods for optimizations.
\end{abstract}





\maketitle

\section{Introduction}
\par Reconstructing 3D human-object interaction (HOI) has numerous applications in fields of robotics, gaming and AR/VR and etc., and monocular methods also have gained growing interest due to the cost-effective advantage. However, as well known, there is ambiguity for depth information from single view, and objects or human bodies can be occluded with high possibility only from single view. These all make the 3D HOI reconstruction from single view still face significant challenges.

\begin{figure}[t!]
    \centering
    \includegraphics[width=\linewidth]{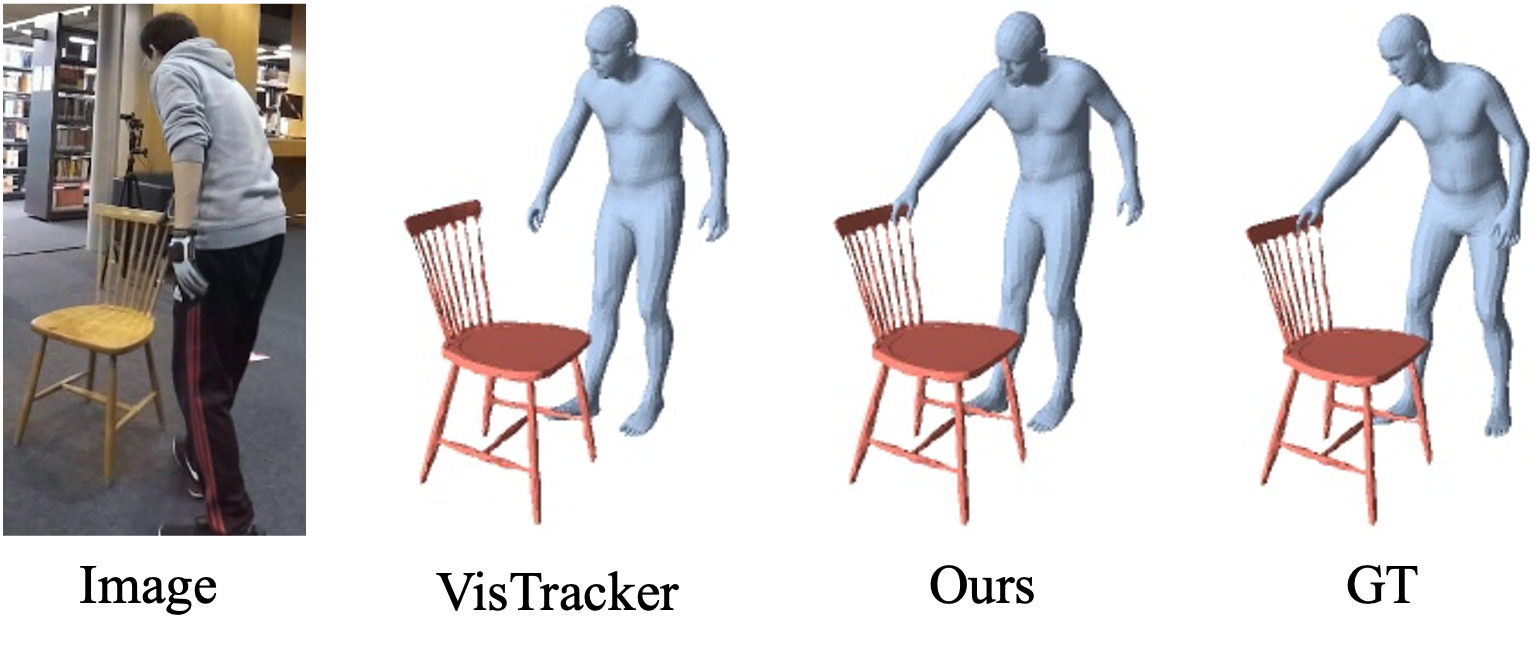}
    \caption{Given a challenging occluded-view image, compared to a current SOTA method VisTracker, our model is able to produce more accurate 3D human body pose and position, outperforming pervious work on the standard benchmark.}
    \label{fig:1}
    \vspace{-0.4cm}
\end{figure}

\par The pipeline for single-view 3D HOI reconstruction typically includes: firstly, creating spatial relationships of humans and objects, such as contact maps \cite{zhang2020perceiving}, implicit distance fields \cite{xie2022chore}, \cite{xie2023visibility}, human-object offsets \cite{huo2023stackflow}, etc.; then, training models on 3D HOI datasets to jointly predict the positions and poses of humans and objects \cite{zhang2020perceiving}, \cite{weng2021holistic}, \cite{xie2022chore}, \cite{xie2023visibility}. These data-driven methods have made significant progress across benchmarks \cite{huang2022intercap}, \cite{bhatnagar2022behave}. Nonetheless, there are still some tricky problems: (1) the 3D HOI training datasets are not universal; (2) the spatial relationship between invisible body parts and objects is difficult to be predicted. Fig. \ref{fig:1} shows a failure by a current SOTA method VisTracker, due to the self-occlusion. Such cases are common in practice but still have not been solved. In the field of 3D human pose estimation, many works introduce explicit cues such as key points or masks as guidance for better performance \cite{choi2020pose2mesh}, \cite{moon2020i2l}, \cite{li2021hybrik}. These motivate us strive to find some explicit cues to guide the correct 3D HOI reconstruction.

\par Contact regions offer reliable cues and can be consistently estimated from monocular images, as demonstrated by numerous studies in 3D hand-object interaction reconstruction \cite{yang2021cpf}, \cite{grady2021contactopt}, \cite{zhao2022stability}, \cite{wang2023interacting}, \cite{hu2024learning}. Similarly, for HOI reconstruction, we believe that contact regions on objects can also be taken as a guidance, while the challenge lies in the strategy of driving body parts to reach contact regions. In fact, this process resembles the end-effector localization task in robotics: given a target position, the end-effector of the robotic arm can be driven to the target position by designing an inverse kinematics (IK) solution to calculate the rotation angles of each joint of arm. Inspired by this, we explore the way of describing the joint, the degrees of freedom, and such like factors relating to real-world robots for the SMPL \cite{loper2023smpl} model appearing in 3D HOI reconstruction.  Correspondingly, we introduce neural network in the IK process and propose a more flexible kinematics solution to bridge the significant gap between SMPL mesh and robots. Experimental results demonstrate that compared to the typical numerical methods used in robotics \cite{coleman1996interior}, our solution is better suited for cases where SMPL’s degrees of freedom are redundant. Furthermore, the proposed kinematics model does not require offline training and its online optimization capability allows it to be conveniently embedded as an optimization module into other methods. Besides, to enhance the practicability, we also propose a Contact Region Network (CRRNet) to determine the contact region that is provided an accurate target for the kinematics model. 

\par We evaluate our method on the BEHAVE benchmark. The experiments demonstrate that our approach can robustly reconstruct the 3D HOI even under occluded views, outperforming the current SOTA method, as has been displayed in Fig. \ref{fig:1}. Our main contributions are summarized as below:

\begin{itemize}
\item We propose a kinematics-based 3D HOI reconstruction method driving the body part represented with SMPL meshes to a target position, which introduce the IK theory and is developed to an online model needing not data training;

\item The Multi-Layer Perceptron (MLP) is introduced into the progress of IK solution, which displays much better performance than the traditional numerical method used in robotics; 

\item We propose a CRRNet to robustly and accurately detect contact regions on objects as the target of the proposed kinematics model, from a monocular video;

\item Our method exhibits good portability and can be simply used for subsequent optimizations in other methods.
\end{itemize}

\section{Related work}

\subsection{Single-view 3D Human-Object Interaction Reconstruction}

\par Due to fast development of deep learning, significant progress has been made in single-view 3D human pose estimation \cite{choi2020pose2mesh}, \cite{moon2020i2l},  \cite{lin2021end}, \cite{li2021hybrik}, \cite{cho2022cross}, \cite{cai2024smpler} and 6D object pose estimation \cite{li2023nerf}, \cite{hai2023shape}, \cite{jiang2024se} separately. However, the accurate estimation of 3D HOI from single view is still challenging only relying on the network training, especially for occlusion cases. Chen et al. \cite{chen2019holistic++} integrate 3D scene reconstruction and 3D human pose estimation, and simplify the representation of object and human body using 3D bounding boxes and skeletons. Zhang et al. \cite{zhang2020perceiving} and Weng et al. \cite{weng2021holistic} further extend this method to reconstruct human and object meshes. They manually define contact regions and jointly optimize object and human body meshes in the scene using heuristic methods. However, these methods suffer from poor scalability and accuracy. 

\par Xie et al. \cite{bhatnagar2022behave} create the first dataset related to HOI, named BEHAVE, and propose a model CHORE \cite{xie2022chore} leveraging implicit neural distance fields to describe the spatial relationships between human and object meshes in single-view images. CHORE is demonstrated impressive performance on the BEHAVE benchmark, but mistakes still easily appear under occluded views. Expanding from CHORE, VisTracker \cite{xie2023visibility} uses a single-view video as input to predict occluded frames by leveraging adjacent frames. The estimation of object poses is improved, while the spatial relationship between humans and objects still remains inaccurate usually. Recently, two works gain impressive results by refining interaction modeling and incorporating object shape priors. Huo et al. \cite{huo2023stackflow} propose to represent the human-object spatial relationships using the offsets of the densely sampled anchor points on human and object mesh surfaces. Mirmohammadi et al. \cite{mirmohammadi2023reconstruction} learn shape priors from a large number of 3D shape datasets and reconstruct object shapes and poses under the constraint of 3D human bodies using 3D bounding boxes as cues. These methods all depend on offline training with extensive 3D HOI data, while currently the HOI dataset is rare, which still leads to large errors unavoidably in the case of occlusion or depth blur.

\par As mentioned above, compared to human bodies, objects are often simplified as rigid bodies, and the estimation of 6D pose and contact regions on objects are more accurate and reliable, particularly in continuous videos. Based on this, we propose to detect contact regions on objects and then drive human body to reach these regions.

\subsection{Methods with Contact Region as Guidance}

\par The contact between human bodies and objects is crucial for guiding tasks like 3D reconstruction, human motion capture, and motion prediction. Contact regions exist simultaneously on the  meshes of human body and object. Some methods, like BSTRO \cite{huang2022capturing} and DECO \cite{tripathi2023deco}, directly predict contact regions on human body meshes from monocular images. However, only guided with contact regions on human body lack reliability since the human body and its movements are complex and ever-changing, while contact regions on rigid objects are often deemed reliable for guiding downstream tasks. Mao et al. \cite{mao2022contact} predict human motion using a distance-based contact graph as guidance, which is created by capturing the contact associations between each human joint and every 3D scene point across time frames. Shimada et al. \cite{shimada2022hulc} propose a human motion capture model HULC guided by dense contact regions estimated between human body parts and static scenes. These methods all rely on additional inputs, the ground-truth point clouds of static scene, thus they are not suitable for the reconstruction of highly dynamic HOI, and in addition, some methods also require inputting multi-view images.

\par It's worth noting that methods with single view input and guided with contact have been extensively explored and applied in 3D hand-object interaction reconstruction \cite{yang2021cpf}, \cite{grady2021contactopt}, \cite{zhao2022stability}, \cite{wang2023interacting}, \cite{hu2024learning}. Yang et al. \cite{yang2021cpf} and Grady et al. \cite{grady2021contactopt} propose the models CPF and ContactOpt, respectively, both using a two-stage strategy that first roughly reconstructs meshes of hand and object and then estimates the contact regions to guide the movement of hand. These methods are very effective and have been widely adopted in subsequent works \cite{liu2023contactgen}, \cite{hu2024learning}, but they cannot be applied to HOI related tasks directly because human body is much more complex than hands and correspondingly the interactions are even more diverse. However, this two-stage strategy is very feasible for tasks with only single view and thus is adopted in this paper.

\begin{figure*}[t!]
    \centering
    \includegraphics[width=\linewidth]{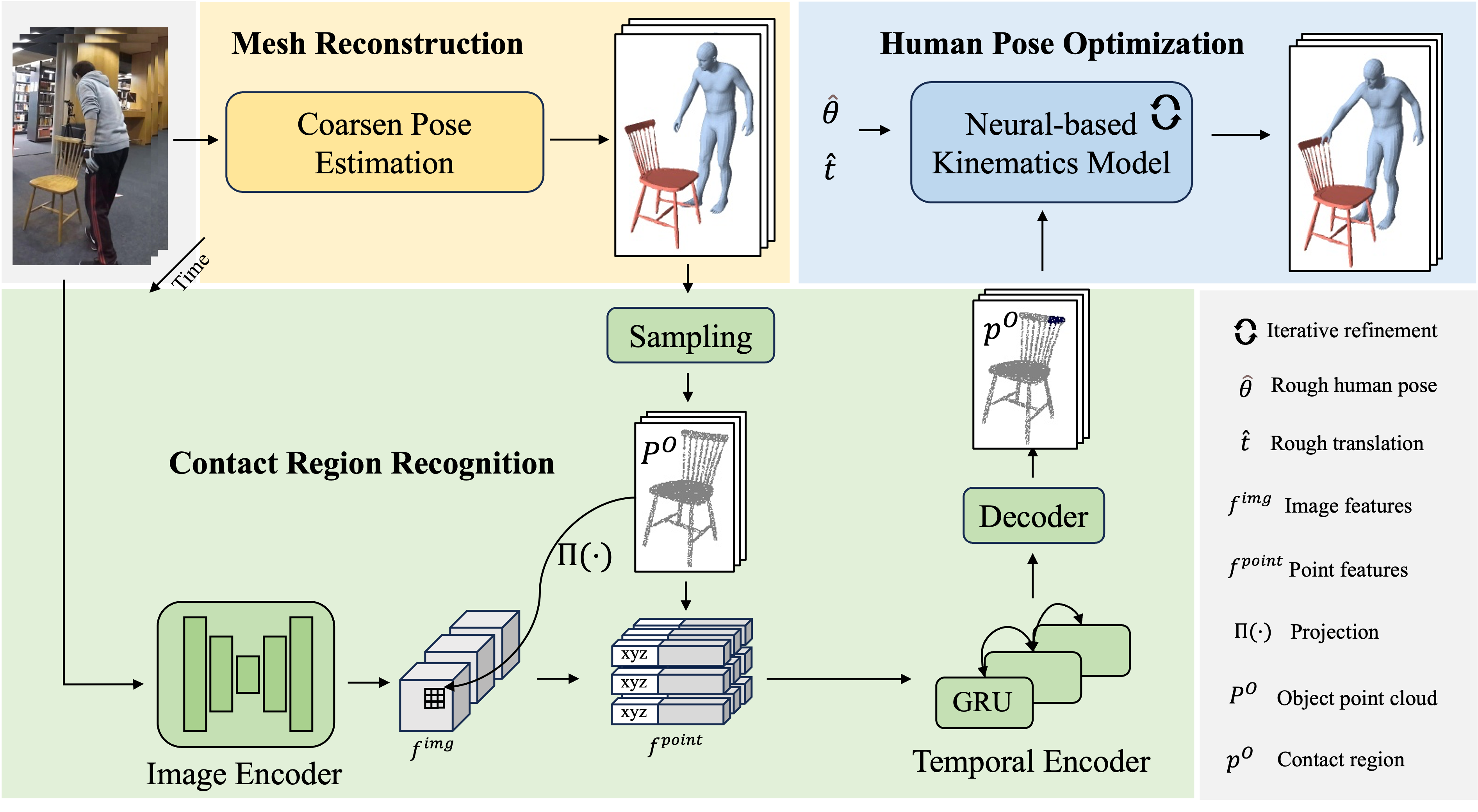}
    \caption{The framework of our proposed method, consisting of: (1) Mesh reconstruction: given a video sequence, the meshes of objects and human bodies are estimated with rough poses by using existing method; (2)Contact Region Recognition: the features of input image sequence and originally estimated point cloud are integrated to estimate the contact regions on the object surface. (3) Human pose optimization: a neural-based kinematics model is proposed to actively guide the human body to reach a contact region.}
    \label{fig:2}
\end{figure*}

\subsection{Kinematics-based Methods Related to 3D Human}

\par There have been some works introduce kinematics to enhance the performance of 3D human pose estimation or human body reconstruction \cite{li2021hybrik}, \cite{li2023niki}, \cite{li2023hybrik}. HybrIK \cite{li2021hybrik} is the most remarkable method that takes the 3D human joint estimated from images as targets and introduces the forward-inverse kinematics to solve for the rotations of corresponding joints in SMPL model. This method largely improves the accuracy of single-view 3D human reconstruction and achieves accurate results comparable to 3D keypoints estimation. However, the kinematics solution in HybrIK highly depends on each previously estimated 3D human joint, and so it cannot be used for 3D HOI reconstruction. 

\par While kinematics-based methods have been explored in human-object interaction synthesis, based on recent progress in auto-regressive model \cite{ghosh2023imos} and diffusion models \cite{pi2023hierarchical}, \cite{li2023object}. However, they often encounter issues like penetration and floating due to limited modeling of dynamics and contacts. Physics-based approaches \cite{pan2023synthesizing}, \cite{hassan2023synthesizing}, \cite{wang2023physhoi} can mitigate these issues but require tailored training strategies for specific motion segments. Recently, Hassan et al. integrate the actions such as lifting, sitting, and lying down into one model, but the model is limited to some HOI scenes and to make it generalizing still remains challenging.

\par In robotics, the angles of each joint can be solved with inverse kinematics, given the end-effector position of a robot. Drawing inspiration from this, we frame the estimation of human pose as an inverse kinematics problem akin to robotic arms' end-effectors, where the contact regions on objects can be defined as the target positions.

\section{Method}
\par The overall architecture of our framework is shown in Fig. \ref{fig:2}. To start with, we outline the task setup and give a concise overview of the forward kinematic (FK) of SMPL; subsequently, our proposed neural-hybrid kinematics model is introduced; finally, the proposed Contact Region Recognition Network is presented.

\subsection{Preliminary}
\textbf{Task setup.} 
As mentioned above, we also conduct a two-stage strategy in this work. The first stage is to create the meshes of objects and human bodies, which have been researched exhaustively. In this work, the object is reconstructed like in \cite{xie2023visibility} with parameters of rotation $\boldsymbol{R}^O\in\mathbb{SO}(3)$ and translation $\boldsymbol{t}^O \in \mathbb{R}^3$. The human is presented by the widely used 3D parametric model SMPL firstly \cite{loper2023smpl}, and then SMPL model is transformed to a human mesh as $\mathcal{M}^S=SMPL(\theta, \beta, \boldsymbol{t}^S)$, with $\beta \in \mathbb{R}^{10}$ for controlling shape, $\theta\in\mathbb{R}^{24 \times 3}$ for representing the rotation of 24 joints, and $\boldsymbol{t}^S\in\mathbb{R}^3$ for the translation.

\textbf{Forward kinematics of SMPL.} The second stage is the kinematics solution for the joints of SMPL, which include the FK and IK algorithms, to calculate a joint coordinate of SMPL and the offsets of the joint targeted with the contact region, respectively. 

With the SMPL model, the rest pose template $\tilde{q}\in\mathbb{R}^{24\times 3}$ can be obtained and the relative rotation matrix $\tilde{\boldsymbol{R}}\in\mathbb{R}^{24\times 3 \times 3}$ can be calculated by the $\theta\in\mathbb{R}^{24\times 3}$. Then the FK solution of SMPL aims to computing the coordinates of the 24 joints, denoted as $\hat{q}\in\mathbb{R}^{24\times 3}$, using the rest pose template $\tilde{q}\in\mathbb{R}^{24\times 3}$ and the relative rotation matrix $\tilde{\boldsymbol{R}}\in\mathbb{R}^{24\times 3 \times 3}$. This is achieved by recursively rotating the template body parts from the root joint to the leaf joint.

Specifically, the parent joint ${pa}(i)$ of root ($i=0$) is obtained first as:
\begin{equation}
    \hat{\boldsymbol{T}}_0=\begin{bmatrix}
  \tilde{\boldsymbol{R}}_0&\tilde{q}_0 \\
  0&1
\end{bmatrix}.
\end{equation}

Then, for the $i^{th}$ joint, the transformation matrix $\hat{\boldsymbol{T}}_i \in \mathbb{R}^{4\times 4}$ to the corresponding root can be computed as:
\begin{equation}
    \hat{\boldsymbol{T}}_i=\hat{\boldsymbol{T}}_{pa(i)}\tilde{\boldsymbol{T}}_i,
\end{equation}
where, $\tilde{\boldsymbol{T}}_i \in \mathbb{R}^{4\times 4}$ denotes the relative transformation:
\begin{equation}
\tilde{\boldsymbol{T}}_i=\begin{bmatrix}
  \tilde{\boldsymbol{R}}_i&\tilde{q}_i-\tilde{q}_{pa(i)} \\
  0&1
\end{bmatrix}. 
\end{equation}
Considering the global translation $\hat{\boldsymbol{t}}$, the coordinate of the $i^{th}$ joint can be calculated as:
\begin{equation}
    \hat{q}_i = \hat{\boldsymbol{T}}_i[:3,3] + \hat{\boldsymbol{t}}
\end{equation}

\subsection{The Proposed Kinematics Model}
If an originally rough SMPL mesh $\hat{\mathcal{M}}^S=SMPL(\hat{\theta},\hat{\beta},\hat{t})$ and a target position $\bar{q}_j\in \mathbb{R}^3$ on an object are given, our method aims to drive the $j^{th}$ joint near the contact region moving from the originally roughly estimated position $\hat{q}_j$ to the target position $\bar{q}_j$, by joints rotation $\Delta \boldsymbol{R}$ and global translation $\Delta \boldsymbol{t}$. As analyzed above, we need to firstly introduce the variables $\Delta \boldsymbol{R}$ and $\Delta \boldsymbol{t}$ into the solution of FK algorithm of SMPL, and then explore the accurate IK solution targeted with contact region on an object for the FK solution. These will be introduced in detail below.

\textbf{The improved forward kinematics for SMPL.} Assume that each body joint has 3 DoFs, according to the twist-and-swing decomposition introducing by Baerlocher et al \cite{baerlocher2001parametrization}, we can decompose the rotation $\Delta \boldsymbol{R}$ into “twist” and “swing”, which further describe the rotation and swing around a joint. For the $i^{th}$ joint, we denote $\phi_i \in \mathbb{R}^1$ and $\boldsymbol{m}_i \in \mathbb{R}^3$ to represent the twist angle and direction, respectively, and denote $\alpha_i \in \mathbb{R}^1$ and $\boldsymbol{n}_i \in \mathbb{R}^3$ the swing angle and direction, respectively.

Then, the $i^{th}$ transformation matrix $\boldsymbol{T}_i$ is rewritten as:

\begin{equation}    \boldsymbol{T}_i=\hat{\boldsymbol{T}}_{pa(i)}\tilde{\boldsymbol{T}}_i\Delta{\boldsymbol{R}}=\hat{\boldsymbol{T}}_{pa(i)}\tilde{\boldsymbol{T}}_i\boldsymbol{T}_i^{tw}\boldsymbol{T}_i^{sw},
    \label{eq:4}
\end{equation}

where $\boldsymbol{T}_i^{tw}$ and $\boldsymbol{T}_i^{sw}$ represent the transformation of twist and swing, respectively. And $\boldsymbol{T}_i^{tw}$ and $\boldsymbol{T}_i^{sw}$ can be written as:

\begin{equation}
    \boldsymbol{T}_i^{tw}=\begin{bmatrix}
  \boldsymbol{R}^{tw}_i & 0\\
0 & 1
\end{bmatrix},
    \boldsymbol{T}_i^{sw}=\begin{bmatrix}
  \boldsymbol{R}^{sw}_i & 0\\
0 & 1
\end{bmatrix},
\label{eq:5}
\end{equation}
where $\boldsymbol{R}^{tw}_i$ and $\boldsymbol{R}^{sw}_i$ represent the rotation of twist and swing, respectively. And $\boldsymbol{R}^{tw}_i$ and $\boldsymbol{R}^{sw}_i$ can be calculated by the Rodrigues formula:
\begin{equation}
    \boldsymbol{R}^{sw}_i=\mathcal{I}+\sin(\alpha_i)[\boldsymbol{n}_i]_{\times}+(1-\cos(\alpha_i))[\boldsymbol{n}_i]_{\times}^{2},
\end{equation}
\begin{equation}
    \boldsymbol{R}^{tw}_i=\mathcal{I}+\sin(\phi_i)[\boldsymbol{m}_i]_{\times}+(1-\cos(\phi_i))[\boldsymbol{m}_i]_{\times}^{2}
\end{equation}
where $[\boldsymbol{n}_i]_{\times}$ and $[\boldsymbol{m}_i]_{\times}$ denote the skew-symmetric matrix of $\boldsymbol{n}_i$ and $\boldsymbol{m}_i$, respectively, and $\mathcal{I}$ is the $3 \times 3$ identity matrix. According to the definition, the twist direction vector $\boldsymbol{m}_i$ can be determined as:
\begin{equation}
    \boldsymbol{m}_i=\frac{\boldsymbol{q}_i}{\left \| \boldsymbol{q}_i \right \|}, \boldsymbol{q}_i=q_i-q_{pa(i)}.
    \label{eq:8}
\end{equation}
Finally, considering the additional translation $\Delta{\boldsymbol{t}}$, the new position of $j^{th}$ joint is:
\begin{equation}
    q_j=\boldsymbol{T}_j[:3, 3]+\hat{\boldsymbol{t}}+\Delta{\boldsymbol{t}}.
\end{equation}
In this improved FK process, the variables $\phi$, $\alpha$, $\boldsymbol{n}$ and $\Delta{\boldsymbol{t}}$ are unknown, which are the goal of the following IK solution.

\begin{figure}[t!]
    \centering
    \includegraphics[width=\linewidth]{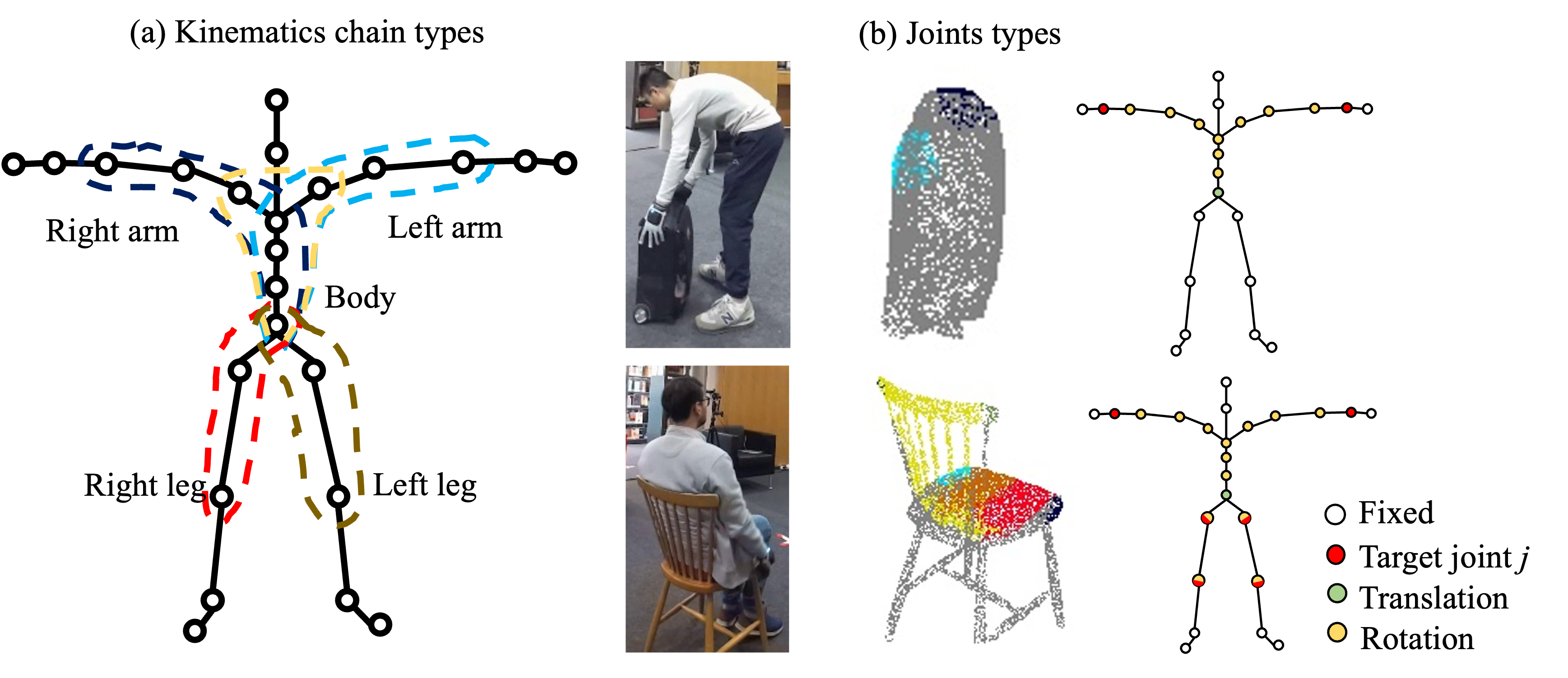}
    \caption{The definition of kinematics chain and joints type. In (a), we define five kinematic chains: left / right arm, left / right leg, body; In (b), we define four joint types: target joint $j$, rotation, translation, fixed. We activate the corresponding kinematic chain based on the type of contact regions and assign the corresponding types to the joints.}
    \label{fig:3}
\end{figure}

\begin{figure}[t!]
    \centering
    \includegraphics[width=\linewidth]{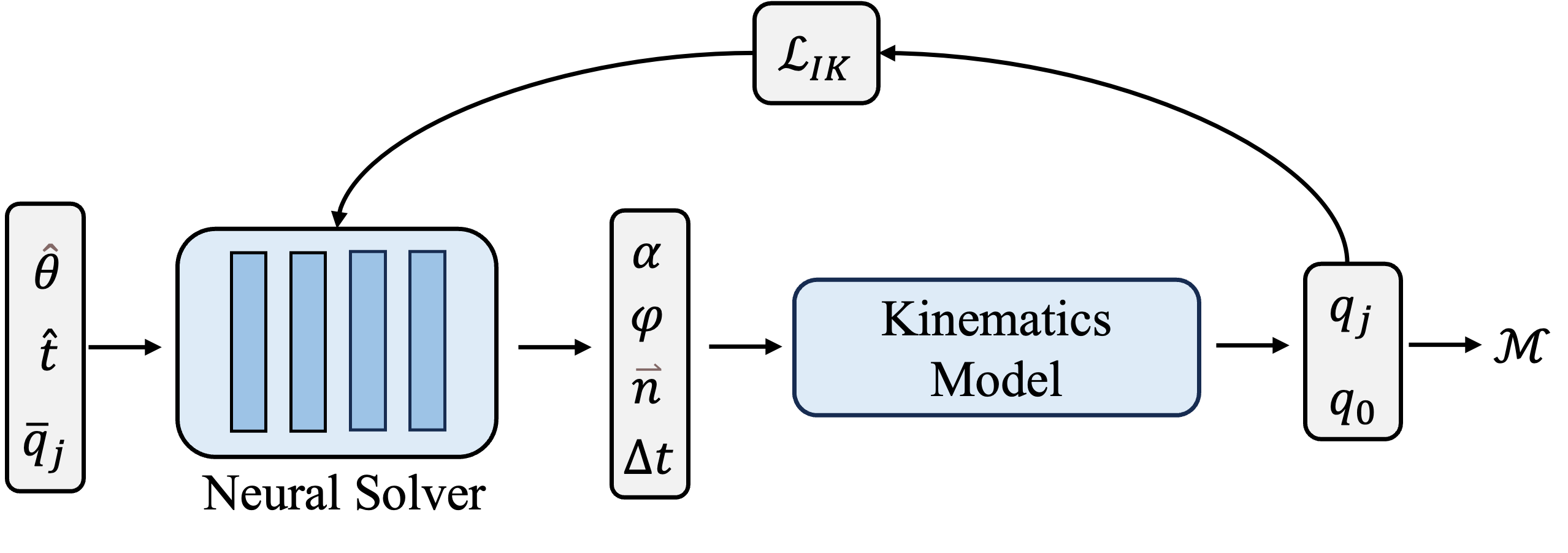}
    \caption{The architecture of our proposed IK solution.}
    \label{fig:4}
    \vspace{-0.6cm}
\end{figure}

\textbf{The inverse kinematics solution based on neural network.} Corresponding to our improved FK algorithm, we have to design an IK algorithm to solve all the unknown parameters. Considering the powerful solving ability of neural network, we design an IK solver based on naive MLP, and formulate a loss function $\mathcal{L}_{IK}$ to supervise the online optimization of the solver parameters, as shown in Fig. \ref{fig:4}.

\begin{figure}[t!]
    \centering
    \includegraphics[width=\linewidth]{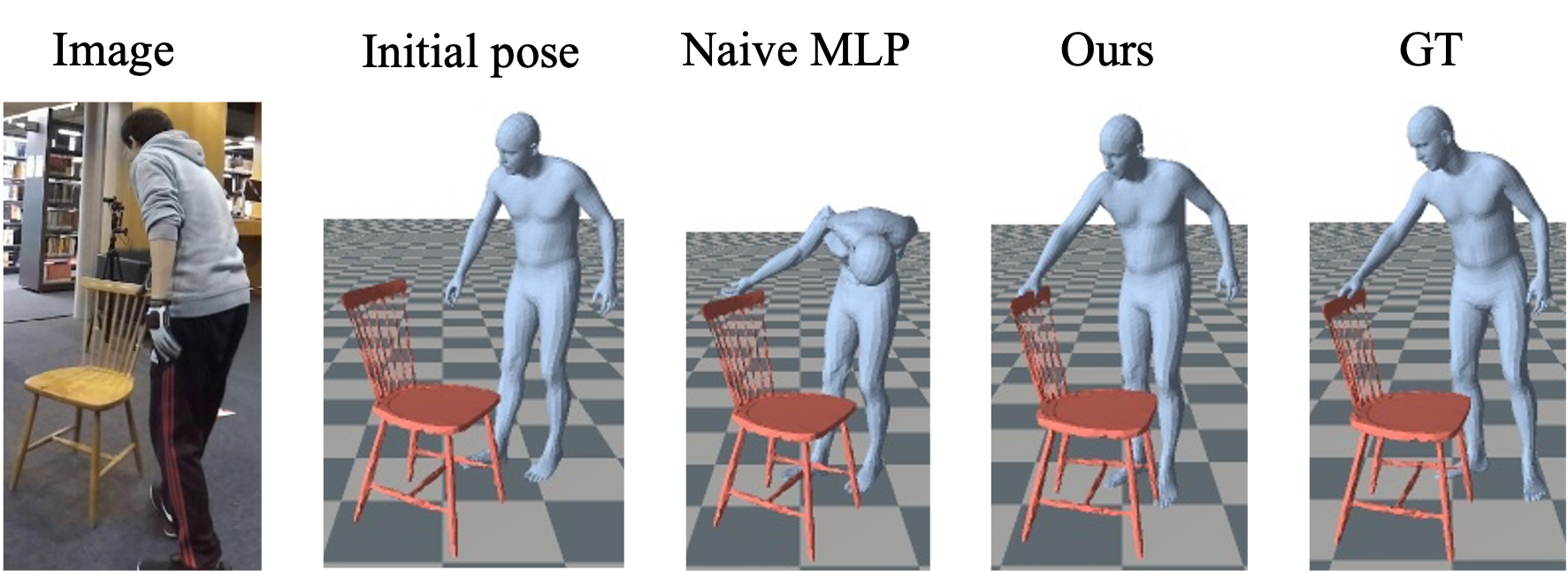}
    \caption{The comparison between our Neural Solver and naive MLP.}
    \label{fig:5}
    \vspace{-0.6cm}
\end{figure}

The inputs of the Neural Solver are the rough pose parameters $\hat{\theta}$, tough global translation $\hat{\boldsymbol{t}}$, and target position $\bar{q}_j$ as inputs, and the outputs are $\phi$, $\alpha$, $\boldsymbol{n}$ and $\Delta{\boldsymbol{t}}$, correspondingly. This process can be expressed as:
\begin{equation}
    (\phi, \alpha, \boldsymbol{n}, \Delta{\boldsymbol{t}})={MLP}(\hat{\theta},\hat{t},\bar{q}_j).
\end{equation}
It is noted that directly predicting $\phi$ and $\alpha$ can lead to peculiar poses, shown in Fig. \ref{fig:5}. To address this issue, we restrict $\alpha$ and $\phi$ to a small range $-\gamma\sim\gamma$ using hyperbolic tangent function (Tanh). Denote the MLP outputs $y_\phi$ and $y_\alpha$, $\sin(\phi)$ and $\cos(\phi)$ can be determined as:
\begin{equation}
    \sin(\phi)=\sin(\gamma)\tanh(y_{\phi}), \cos(\phi)=\sqrt{1-\sin(\phi)^2},
\end{equation}
\begin{equation}
    \tanh(y_\phi)=\frac{e^{y_\phi}-e^{-y_\phi}}{e^{y_\phi}+e^{-y_\phi}}
\end{equation}

The same operation is applied in predicting $y_\alpha$. The results are shown in Fig. \ref{fig:5}.

In addition, due to the large number of redundant DoFs in SMPL kinematics model, there are theoretically countless sets of solutions. To supervise the solver find the optimal solution, we 
further introduce 2D root joint loss $\mathcal{L}_{2D}=(q_0^{2D}-\bar{q}_0^{2D})^2$, where $\bar{q}_0^{2D}$ is obtained from PoseNet \cite{moon2019camera}. $\mathcal{L}_{2D}$ helps for ensuring that the mesh is in an accurate global translation. Thus, $\mathcal{L}_{IK}$ can be written as:
\begin{equation}
    \mathcal{L}_{IK}=\epsilon_1(q_j-\bar{q}_j)^2+\epsilon_2(q_0^{2D}-\bar{q}_0^{2D})^2,
\end{equation}
where $\epsilon_1$ and $\epsilon_2$ are hyper parameters. Compute the initial value $\xi$ of $\mathcal{L}_{IK}$, and optimize the parameters of Neural Solver. When $\mathcal{L}_{IK}<0.01\xi$, the iteration can be stopped and output the final result.

\textbf{Implementation details. } In practice, since we rely solely on contact regions as the guidance, we cannot optimize all joints simultaneously. The abundance of redundant DoFs can lead to failure in IK solving. Our approach is to activate the corresponding kinematic chain based on the contact body part, and for joints not on the chain, we keep them fixed. As shown in Fig. \ref{fig:3}.

Besides, in modeling, we assume $\bar{q}_j$ as the coordinates of a single point, but in fact, $\bar{q}_j$ should be the coordinates of a set of points in the contact regions. So, we apply average distance between $q_j$ and $\bar{q}_j$ to supervise the IK process.

\subsection{Contact Region Recognition Network}

Since none of existing 3D HOI datasets have ground-truth annotations for contact regions, for training and testing CRRNet, we first design a solution to generate contact region pseudo-labels from the GT human-object mesh. Fig. \ref{fig:2} illustrates the overall architecture of our proposed CRRNet. The structure of CRRNet is mainly based on PiCR in CPF \cite{yang2021cpf}. To improve the recognition stability, we design a feature fusion module and a temporal encoder.

\textbf{Contact regions generation.} In this paper, the contact regions on objects are represented as sets of points. In our definition, each point on objects contains four attributes, among where three of them are points’ coordinates, and another is the contact type. According to whether a point on the object is close enough to human body and which body part is the closest, the points on the object are categorized into 15 contact types. The first 14 types correspond to the 14 parts of the SMPL model vertices, as illustrated in Figure 3, while the $15^{th}$ type represents no-contact.

\begin{figure}[t!]
    \centering
    \includegraphics[width=\linewidth]{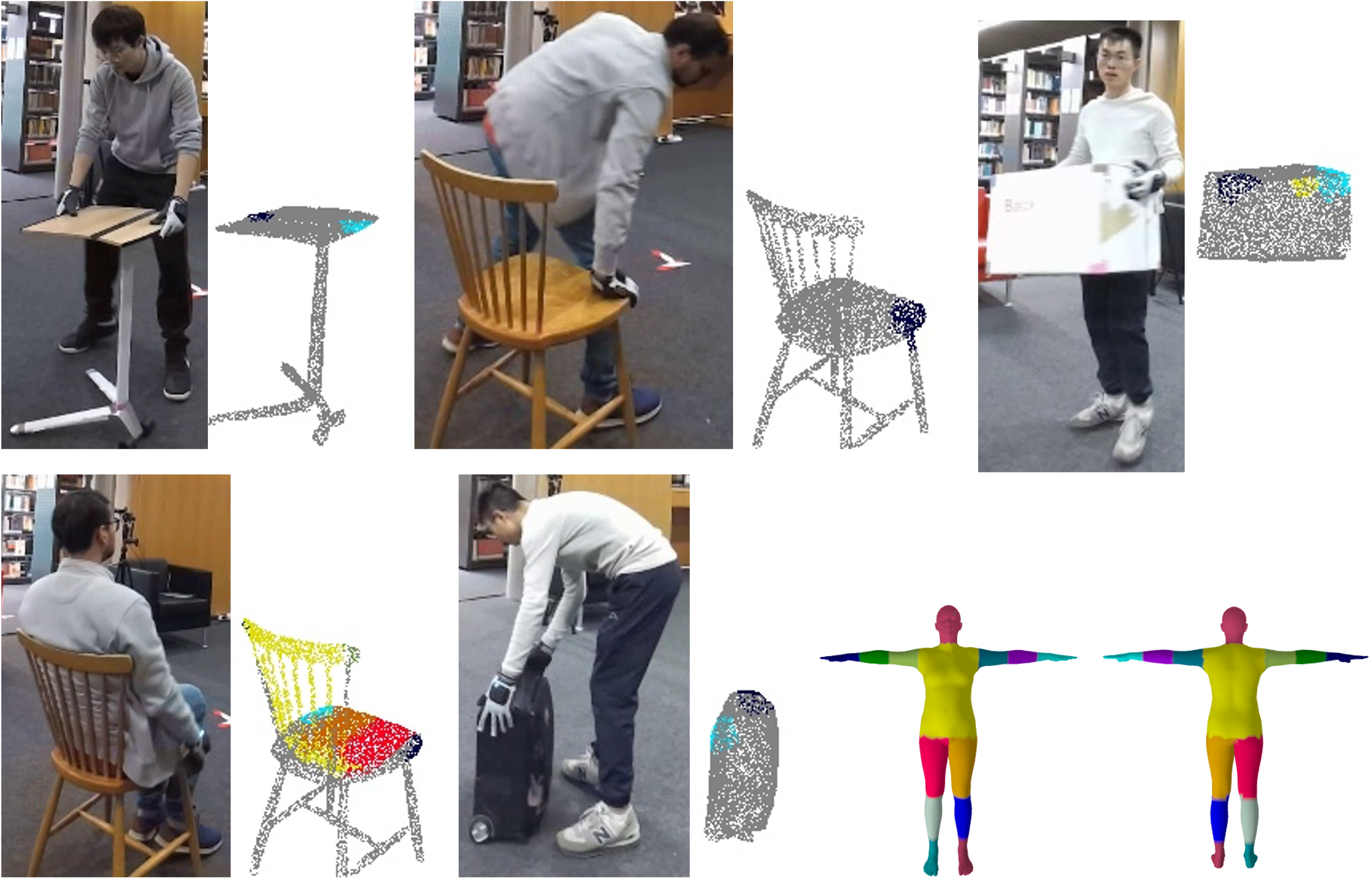}
    \caption{Some results of contact regions generation.}
    \label{fig:6}
    \vspace{-0.6cm}
\end{figure}

Let $P^O\in\mathbb{R}^{N_O\times 3}$ and $P^S\in\mathbb{R}^{N_S\times 3}$ denote the coordinates, one-hot vectors ${p^O\in\mathbb{R}^{N_O\times 15}}$ and ${p^S\in\mathbb{R}^{N_S\times 14}}$ denote the types of the object point clouds and SMPL mesh vertices. For the $i^{th}$ point, we compute out it's contact type $p_i^O$ by two steps: (1) query the nearest point $P_j^S$ on the human mesh, and the distance between $P_j^S$ and $P_j^O$ is:
\begin{equation}
    d_i=\min_{j\in [0,N_O)}\left \| P_j^O-P_j^S \right \|_2.
\end{equation}
(2) Set the threshold of $d_i$ to 0.04, compute the $p_i^O$:
\begin{equation}
    p_i^O=\begin{cases}
 p_j^S, d_i<0.04\\
[0, \dots, 0, 1], d_i \ge 0.04
\end{cases}.
\end{equation}
Some results of contact regions generation are shown in Fig. \ref{fig:6}.

\textbf{Image-point feature fusion.} Given an video ${\left \{ I_\tau \right \}}_{\tau=1}^T$, where $I_{\tau}\in\mathbb{R}^{H\times W \times 3}$ with $H$ and $W$ representing the height and width of the image, respectively. The image encoder is designed based on Hourglass \cite{newell2016stacked}. The image encoder sequentially takes $I_{\tau}$ as input and extracts a sequence of feature maps $\left \{ f_\tau^{img} \right \}_{\tau=1}^T$, where $f_\tau^{img} \in \mathbb{R}^{{\frac{H}{4}}\times\frac{W}{4}\times C_{img}}$, and $C_{img}$ denotes the number of channels in the feature map. Correspondingly, the sequence of object point cloud coordinates is denoted as $\left \{ P_\tau^O \right \}_{\tau=1}^T$, where $P_\tau^O\in\mathbb{R}^{N^O\times 3}$. 

In the feature fusion stage, for the $i^{th}$ point at time $\tau$,
it is first projected onto the pixel space $(P_{\tau,i}^O)_{2D}=\Pi(P_{\tau,i}^O)$, and then the corresponding image features $f_{\tau,i}^{img}=f_{\tau}^{img}|_{(P_{\tau,i}^O)_{2D}}$  and point cloud coordinates $P_{\tau,i}^O$ are fused together to form the point features $f_{\tau,i}^{point}=[P_{\tau,i}^O,f_{\tau,i}^{img}]\in\mathbb{R}^{3+C_{img}}$. However, considering the errors between the input point cloud and the GT, incorrect feature positions on the image plane may lead to unstable recognition.

In this paper, we propose to comprehensively consider all features within the $k \times k$ neighborhood around $(P_{\tau,i}^O)_{2D}$ (where window size $k$ is an odd number), and take the average of these features as the representative feature of the point. $f_{\tau,i}^{img}$ can be rewritten as:
\begin{equation}
    f_{\tau,i}^{img}=\frac{1}{k\times k}\sum_{i=0}^{k-1}\sum_{j=1}^{k-1}f_{\tau}^{img}|_{(P_{\tau,i}^O)_{2D}+(i-\frac{k-1}{2},j-\frac{k-1}{2})}.
\end{equation}

\begin{figure*}[t!]
    \centering
    \includegraphics[width=\linewidth]{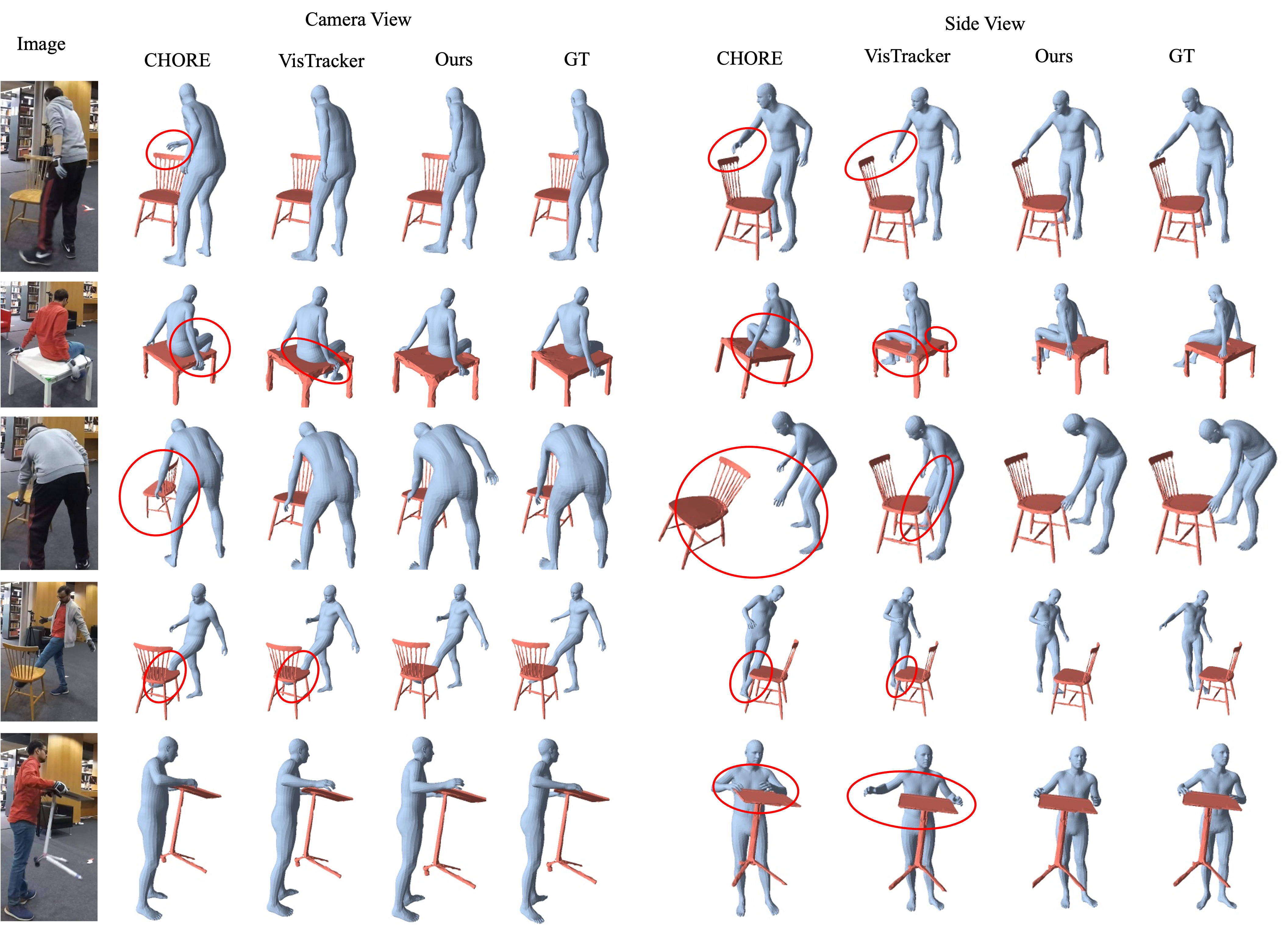}
    \caption{Comparison with CHORE \cite{xie2022chore}, VisTrakcer \cite{xie2023visibility} on BEHAVE \cite{bhatnagar2022behave}. The red circles mark the inaccurate parts. CHORE consistently fails to accurately estimate the distance between the human and the object. VisTracker, on the other hand, integrates spatial and temporal information, resulting in more precise distance estimations compared to CHORE. However, in certain occluded viewpoints, VisTracker may still exhibit biased judgments. In contrast, our method demonstrates superior performance in estimating the distance between the human and the object. Additionally, our method tends to provide more accurate human pose estimations.}
    \label{fig:7}
    \vspace{-0.3cm}
\end{figure*}

\textbf{Temporal encoder.} Existing literature has pointed out that the prediction results of the current frame can benefit from information in past and future frames, especially when the prediction target of the current frame is partially occluded. Sequential architectures can effectively extract information from past and future frames and successfully predict the target of the current frame. In human-object interaction scenarios, occlusion of the contact region is often severe. To address this issue, this paper adopts Gated Recurrent Units (GRU) as the point cloud feature encoder, jointly considering all point cloud features within the sequence. For the $i^{th}$ point at $\tau$, the features after temporal encoder can be represented as $\bar{f}_{\tau,i}^{point}=g(\left\{f_{\tau,i}^{point}\right\}_{\tau=1}^T)$, where $g$ denotes the GRU module.

\textbf{Training procedure.} We train CRRNet based on BEHAVE dataset. Let $\left\{ p_\tau^O\right\}_{\tau=1}^T$ denotes the prediction of CRRNet, where $p_\tau^O\in\mathbb{R}^{N_O\times 15}$, and the GT is denoted as $\left\{ \bar{p}_\tau^O\right\}_{\tau=1}^T$. The training procedure is supervised by cross-entropy loss:
\begin{equation}
    \mathcal{L}_{CRR}=\frac{\epsilon_0}{T}\sum_{\tau=1}^{T}\sum_{i=1}^{N_O}\sum_{j=1}^{15}(-\bar{p}_{\tau,i,j}^O\log(p_{\tau,i,j}^O)),
\end{equation}
where $j$ represents the $j^{th}$ type contact region, and the hyper parameters $\epsilon_0$ is set to 0.006. Adam optimizer is chosen with a learning rate 0.001, and the batch size is set to 8 per GPU. All training tasks are completed on two Nvidia RTX A6000 GPUs.

\section{Experiments}
\subsection{Dateset}

This paper conducts training and evaluation on the indoor dataset BEHAVE \cite{bhatnagar2022behave}. BEHAVE consists of 4 camera views capturing 7 subjects interacting with 20 different objects indoors, annotated with SMPL and object registration information at 1fps. During the training of CRRNet, we follow the official split, with 217 sequences used for training and 82 sequences used for testing. The proposed SMPL pose optimization model in this paper does not require offline training, thus only the 82 sequences are used for testing. For fairness, other methods in this paper are also tested on the same 82 sequences following the official split.

\subsection{Evaluation metrics}

Previous methods such as CHORE and VisTracker typically perform Procrustes Alignment on the SMPL mesh before computing the error, which does not reflect the absolute deviation of the SMPL mesh. Therefore, we use two types of Chamfer distances to assess the absolute deviation and the relative deviation of the human body after Procrustes Alignment, denoted as \textit{Chamfer} and \textit{PA-Chamfer}, respectively, measured in centimeters.

\subsection{Comparison with state-of-the-art}

We compare our human reconstruction results against the baseline methods PHOSA, CHORE, and VisTracker, reporting \textit{PA-Chamfer} and \textit{Chamfer} errors in Tab. \ref{table:1}. Compared to the three methods, our method achieves the lowest \textit{PA-Chamfer}, demonstrating more accurate estimation of human body pose. Similarly, our method also achieves the lowest \textit{Chamfer}, indicating more accurate positioning of the human body in space. Additionally, qualitative comparison results are shown in Fig. \ref{fig:7}.

\begin{table}[h]
\caption{Comparison with SOTA methods.}
\vspace{-0.3cm}
\label{table:1}
\begin{center}
\begin{tabular}{ccc}
\hline
Method & $\textit{Chamfer}\downarrow$ & $\textit{PA-Chamfer}\downarrow$  \\
\hline
PHOSA & - & 12.86 \\
CHORE & 16.84 & 5.55 \\
VisTracker & 13.16 & 5.25 \\
\hline
Ours & \bf11.42 & \bf5.12 \\
\hline
\end{tabular}
\end{center}
\vspace{-0.5cm}
\end{table}

\subsection{Ablation study on Neural-hybrid Kinematics Model}

Solving SMPL IK is an ill-posed problem, and to obtain the optimal solution, there are three key designs in the neural-hybrid kinematics model: MLP-based neural solver, 2D loss $\mathcal{L}_{2D}$, restricting $\phi$ and $\alpha$ to $-\gamma\sim\gamma$. In this section, we design experiments to prove their effectiveness, respectively. Besides, we design another experiment to explore the impact of the value of $\gamma$ on the results.

\begin{figure}[t!]
    \centering
    \includegraphics[width=\linewidth]{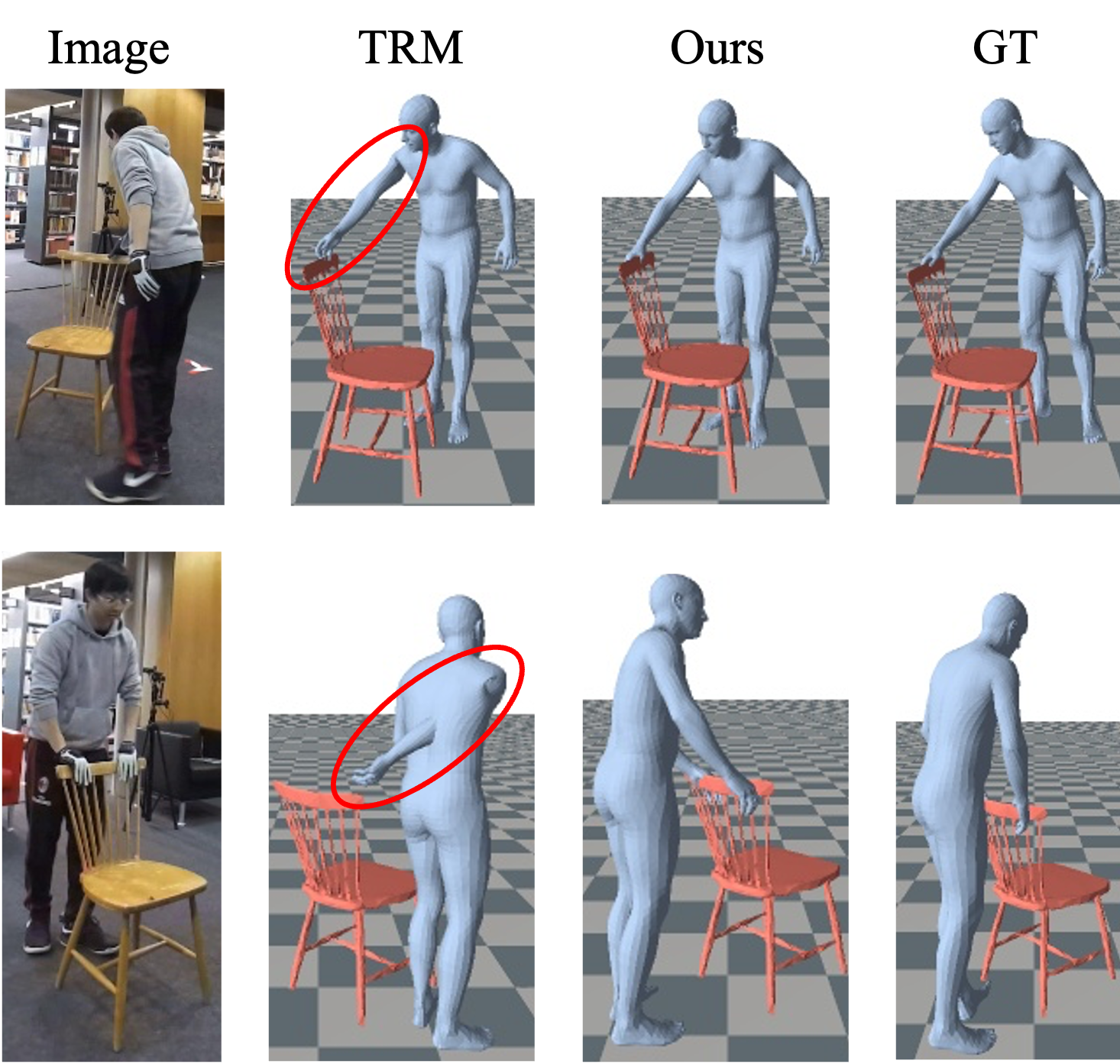}
    \caption{The comparison between our proposed Neural Solver and Trust Region Method (TRM) \cite{coleman1996interior}.}
    \label{fig:8}
    \vspace{-0.6cm}
\end{figure}

\textbf{Effectiveness of Neural Solver.} In robotics, each joint of a robotic arm is typically simplified to have only 1-2 DoFs, and the entire arm usually does not exceed 7 DoFs. In such cases, traditional numerical methods can yield very accurate IK solutions. However, the SMPL model has 24 joints, each with 3 DoFs. Considering the arm kinematic chain starting from the root joint, even without considering the wrist joint, there are still 7 joints, resulting in a total of 21 DoFs. For such highly redundant DoFs, traditional numerical methods are inadequate for solving the inverse kinematics of the SMPL model. To substantiate this assertion, we apply the widely-used Trust Region Method (TRM) \cite{coleman1996interior} to tackle the joint rotations of SMPL model. The results are shown if Fig. \ref{fig:8}. Although the specified joint can reach the target position, the other joints tend to exhibit peculiar rotations. In contrast, neural-based methods are more flexible and allow us to customize more objective function constraints for the network, leading to better results. As shown in Fig. \ref{fig:8}, the poses solved using Neural Solver are more accurate and natural.

\begin{figure}[t!]
    \centering
    \includegraphics[width=\linewidth]{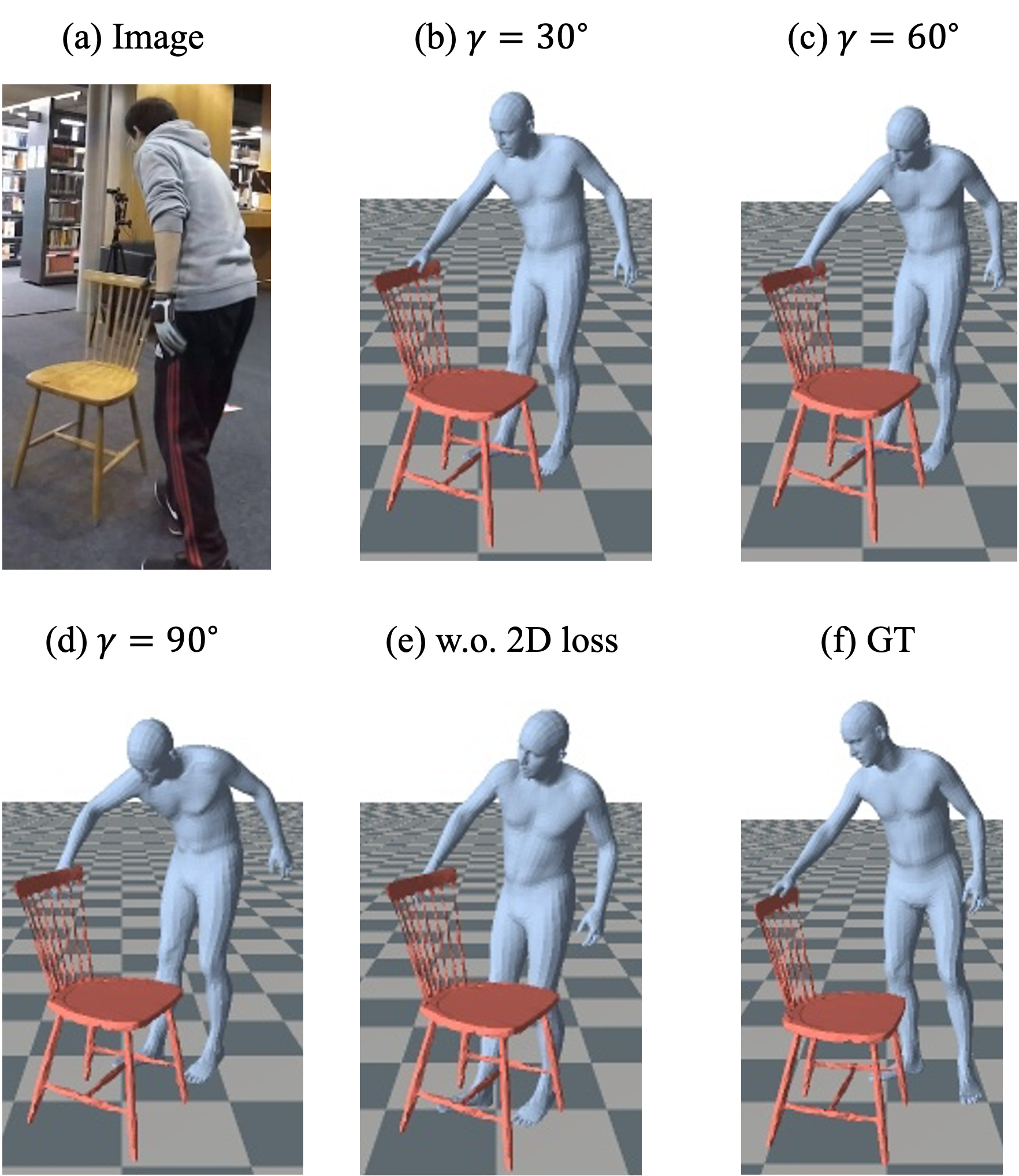}
    \caption{The results of ablation study on $\mathcal{L}_{2D}$ and $\gamma$.}
    \label{fig:9}
    \vspace{-0.6cm}
\end{figure}

\textbf{Effectiveness of $\mathcal{L}_{2D}$.} In the IK process, due to the significant redundancy in degrees of freedom, theoretically, there may be no feasible solution if only relying on fitting model parameters constrained by target positions. The typical approach is to add constraints. However, since we cannot obtain the 3D coordinates of the joints, it is challenging to constrain the positions of all joints on the kinematic chain. Our approach is to utilize two-dimensional keypoints as guidance. Compared to 3D joints estimation, existing 2D keypoints detectors are highly reliable. Keeping $\gamma={30}^{\circ}$, we remove the 2D loss for solving, and the results, as shown in Fig. \ref{fig:9} (d), indicate that the distance error between the human body and the object increases. Additionally, due to the incorrect positioning of the root joint of the human body, the arm posture is also inaccurate. When reintroducing the 2D loss, the reconstruction results, as shown in Fig. \ref{fig:9} (b), exhibit improvement. We repeated our experiments on multiple images and consistently obtained the same results.

\textbf{The impact of $\gamma$.} To avoid generating exaggerated poses, we limit the solution range of $\phi$ and $\alpha$ to $-\gamma\sim\gamma$. Li et al. \cite{li2021hybrik} study the distribution of twist angle $\phi$ in their work HybrIK, and experimental results show that only a few joints have a range over ${30}^{\circ}$, while other joints have a limited range of twist angle. Unlike HybrIK, which directly generates rotations from the rest pose, out method starts from an initial rough pose. Therefore, the range of $\phi$ should be less than ${30}^{\circ}$, while the range of $\alpha$ cannot be determined. Based on above analysis, we conduct experiments with $\gamma$ set to ${30}^{\circ}$, ${60}^{\circ}$, ${90}^{\circ}$, respectively. The experimental results are shown in the Fig. \ref{fig:9}. When $\alpha$ and $\phi$ are restricted within ${30}^{\circ}$, our method can successfully drive the twist to the target, and also estimate an accurate pose. However, when $\gamma$ is increase to ${60}^{\circ}$ and ${90}^{\circ}$, although our method can still drive the twist to the target, it’s obvious that other joints along the right arm produce redundant rotations, resulting in the larger pose error, as illustrated in Tab. \ref{table:2}.

\begin{table}[h]
\caption{The impact of $\gamma$.}
\vspace{-0.3cm}
\label{table:2}
\begin{center}
\begin{tabular}{ccc}
\hline
$\gamma$ & $\textit{Chamfer}\downarrow$ & $\textit{PA-Chamfer}\downarrow$  \\
\hline
${30}^{\circ}$ & \bf11.42 & \bf5.12 \\
${60}^{\circ}$ & 12.67 & 6.31 \\
${90}^{\circ}$ & 13.27 & 6.51 \\
\hline
\end{tabular}
\end{center}
\vspace{-0.5cm}
\end{table}

\subsection{Ablation study on CRRNet}

To enable CRRNet to reliably recognize contact regions from video sequences, two key designs are employed: image-point feature fusion and temporal encoder. The former aims to mitigate the impact of point projection errors, while the latter, by enhancing the connections between frames in a sequence, improves the stability of recognition.

\begin{figure}[t!]
    \centering
    \includegraphics[width=\linewidth]{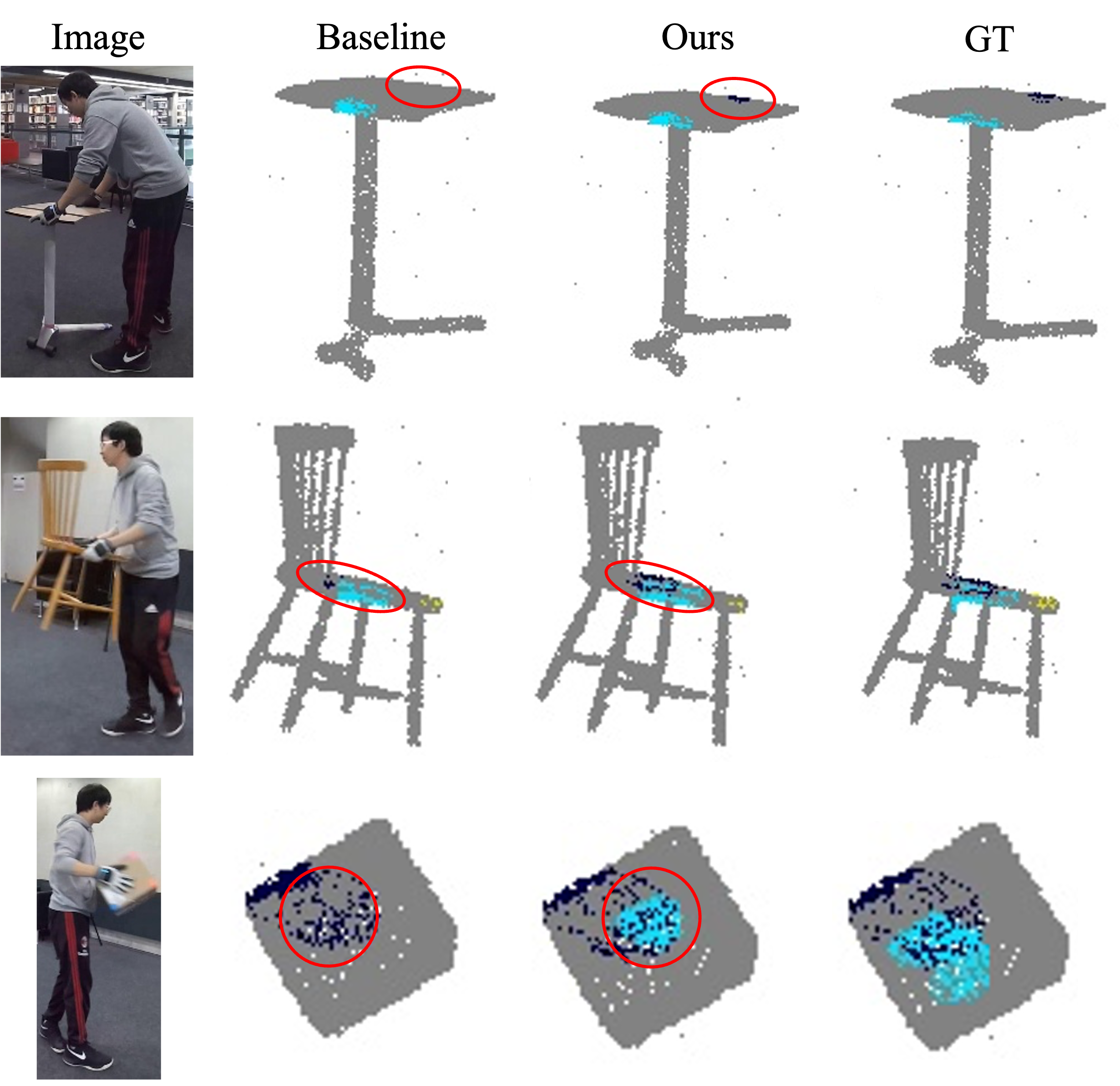}
    \caption{The comparison between CRRNet and baseline.}
    \label{fig:10}
\end{figure}

\textbf{Effectiveness of CRRNet.} In our experiments, we select PiCR as the baseline. To comparison between the baseline and CRRNet is illustrated in Fig. \ref{fig:10}. CRRNet demonstrates superior recognition performance compared to the baseline, particularly in cases where the contact regions are completely occluded, CRRNet is still able to recognize the contact regions.

\begin{figure}[t!]
    \centering
    \includegraphics[width=\linewidth]{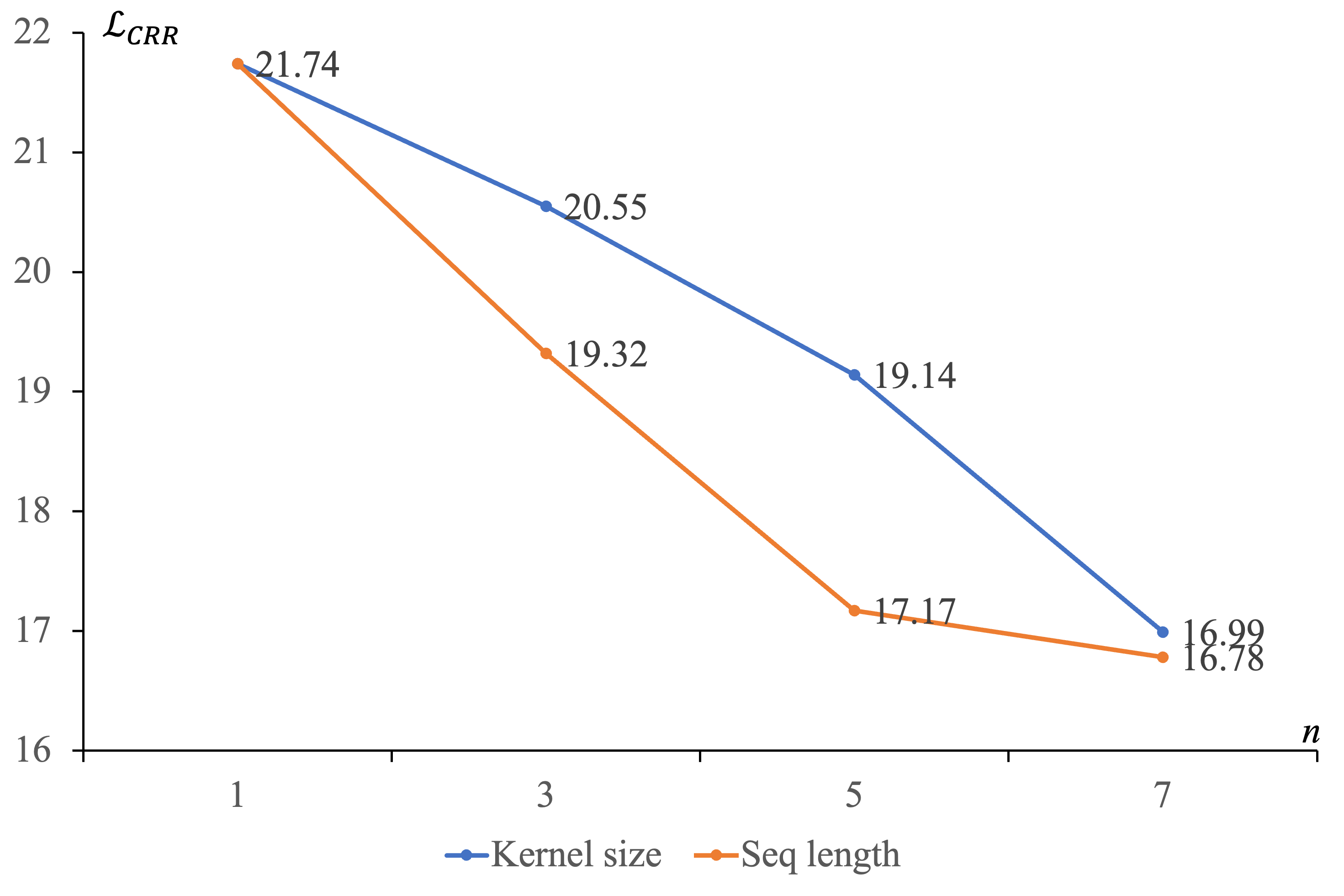}
    \caption{The impact of $k$ and $T$ on $\mathcal{L}_{CRR}$.}
    \label{fig:11}
    \vspace{-0.3cm}
\end{figure}

\textbf{The impact of $k$ and $T$.} In the image-point feature fusion and temporal encoder, the window size $k$ and sequence length $T$ are critical parameters that affect the performance of CRRNet.  Considering both efficiency and memory consumption of CRRNet, we conduct experiments with $k=1,3,5,7$ and $T=1,3,5,7$ respectively, using $\mathbf{L}_{CRR}$ as the evaluation metric, to investigate the impact of $k$ and $T$ on CRRNet's performance. The experimental results are shown in Fig. \ref{fig:11}. As $k$ and $T$ increase, CRRNet's recognition of contact regions becomes more accurate. In practice, we set $k=7$ and $T=7$.

\section{Conclusion}

In this work, we present a kinematics-based framework for reconstructing 3D human-object interaction from a single-view video. We design a neural-hybrid kinematics model to enhance the human-object interactions by driving human body to actively interact with objects. To find the optimal solution in the IK process, we combine MLP and design a stable, accurate, and flexible solver. Additionally, we propose a Contact Region Recognition Network to provide accurate targets for the kinematics model. Through extensive experiments on popular benchmark BEHAVE, our proposed method outperforms the state-of-the-art. In the future work, we plan to extend this work to explore reconstructing interactions between multi-humans and multi-objects.

\newpage


\bibliographystyle{ACM-Reference-Format}
\bibliography{sample-base}

\end{document}